\documentclass[10pt,twocolumn,letterpaper]{article}

\usepackage{cvpr}              

\usepackage{stfloats}

\usepackage{overpic}
\usepackage{bm}
\usepackage{tabu}
\usepackage{tabularx}
\usepackage{bbding}
\usepackage{multirow}
\usepackage{scalerel,stackengine}
\usepackage[accsupp]{axessibility}
\usepackage{array}
\usepackage[linesnumbered,ruled,vlined]{algorithm2e}
\usepackage{algorithmic}

\usepackage[utf8]{inputenc} 
\usepackage[T1]{fontenc}    
\usepackage{url}            
\usepackage{amsfonts}     
\usepackage{nicefrac}     
\usepackage{microtype}      
\usepackage{xcolor}         

\definecolor{cvprblue}{rgb}{0.21,0.49,0.74} 
\usepackage[pagebackref,breaklinks,colorlinks,allcolors=cvprblue]{hyperref} 

\def\paperID{45509} 
\def\confName{CVPR}
\def\confYear{2026}

\title{FlashCap: Millisecond-Accurate Human Motion Capture via Flashing LEDs and Event-Based Vision}

\author{
Zekai Wu$^{1,2}$\textsuperscript{*},
Shuqi Fan$^{1,2}$\textsuperscript{*},
Mengyin Liu$^{1,2}$,
Yuhua Luo$^{1,2}$,
Xincheng Lin$^{1,2}$,\\
Ming Yan$^{1,2,3}$,
Junhao Wu$^{1,2}$,
Xiuhong Lin$^{1,2}$,
Yuexin Ma$^{4}$,\\
Chenglu Wen$^{1,2}$,
Lan Xu$^{4}$,
Siqi Shen$^{1,2}$\textsuperscript{$\dagger$},
Cheng Wang$^{1,2}$\\
$^{1}$Fujian Key Laboratory of Urban Intelligent Sensing and Computing, Xiamen University\\
$^{2}$Key Laboratory of Multimedia Trusted Perception and Efficient Computing,\\ Ministry of Education of China, School of Informatics, Xiamen University\\
$^{3}$National Institute for Data Science in Health and Medicine, Xiamen University\\
$^{4}$ShanghaiTech University
}

\begin{document}

\twocolumn[{%
\renewcommand\twocolumn[1][!htb]{#1}%
\maketitle
\paragraph{\vspace{-14mm}}
\begin{center}
    \centering
    \includegraphics[width=1\linewidth]{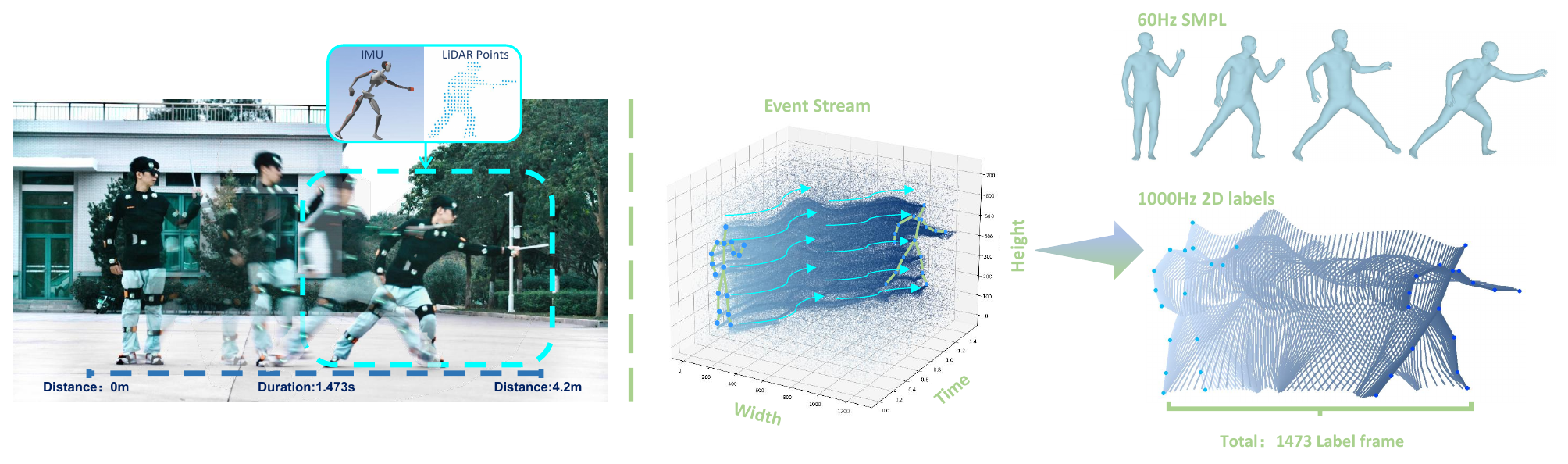}
    \vspace{-7mm}
    \captionof{figure}{FlashCap Overview. Left/Middle: A fencing lunge recorded via our multi-modal, event-based system using flashing LEDs. Right: Generated annotations featuring 1000Hz 2D labels (bottom) alongside 60Hz 3D SMPL (top), capturing fine-grained motion dynamics.}
    
    \label{fig:overview}
\end{center}%
}]

\begingroup
    \renewcommand\thefootnote{\fnsymbol{footnote}}
    \footnotetext[1]{Equal contribution.}
    \footnotetext[2]{Corresponding author.}
 \endgroup
\begin{abstract}
Precise motion timing (PMT) is crucial for swift motion analysis. A millisecond difference may determine victory or defeat in sports competitions. Despite substantial progress in human pose estimation (HPE), PMT remains largely overlooked by the HPE community due to the limited availability of high-temporal-resolution labeled datasets. Today, PMT is achieved using high-speed RGB cameras in specialized scenarios such as the Olympic Games; however, their high costs, light sensitivity, bandwidth, and computational complexity limit their feasibility for daily use. 
We developed \textbf{FlashCap}, the first flashing LED-based MoCap system for PMT. With FlashCap, we collect a millisecond-resolution human motion dataset, \textbf{FlashMotion}, comprising the event, RGB, LiDAR, and IMU modalities, and demonstrate its high quality through rigorous validation. To evaluate the merits of FlashMotion, we perform two tasks: precise motion timing and high-temporal-resolution HPE. For these tasks, we propose \textbf{ResPose}, a simple yet effective baseline that learns residual poses based on events and RGBs. Experimental results show that ResPose reduces pose estimation errors by $\sim$40\% and achieves millisecond-level timing accuracy, enabling new research opportunities. The dataset and code will be shared with the community.
\end{abstract}
\section{Introduction}
\label{sec:intro}

Precise motion timing (PMT) plays a critical role in understanding human motion. In competitive sports, such as running~\cite{Sportskeeda} and speed climbing~\cite{LatitudeClimbing}, millisecond-level temporal discrepancies can determine athletic outcomes. A documented case in Luge competitions demonstrates that a 2-millisecond difference leads to a loss of a bronze medal~\cite{Luge}. It is not uncommon to record human motion with millisecond accuracy. For example, in a speed climbing competition~\cite{SpeedClimbing}, the fastest time to finish the competition for women is 6.447s. Millisecond-level time precision guarantees the integrity of sports results and enables in-depth motion analysis.




Precise temporal analysis of human motion remains understudied in human pose estimation (HPE) research, due to insufficient high-frequency annotated datasets. Current public motion datasets~\cite{wang2024continuous} peak at 120Hz - inadequate for developing millisecond-accurate algorithms required by ultra-fast motion analysis (e.g., swing trajectory reconstruction). 
Existing MoCap systems face fundamental limitations: IMU-based solutions~\cite{Tip} offer 60-240Hz while optical systems~\cite{kuhne2024} achieve up to 330Hz, constrained by hardware sampling capabilities. Conventional RGB cameras (30-60Hz ~\cite{UIP}) lack temporal resolution. Although high-speed RGB cameras ($\geq 1000$ Hz) used in elite sports~\cite{spikegs} offer higher frame rates, their adoption is hindered by lighting, setup requirements, bandwidth/storage demands~\cite{natureEvent}, and cost.

We aim to capture 1000Hz human motion with ground-truth (GT) labels using event cameras, leveraging their high temporal resolution and low bandwidth. Obtaining such high-frequency GT is challenging: existing methods either face hardware sampling ceilings of auxiliary modalities~\cite{relid11d,calabrese2019dhp19} or suffer interpolation errors during swift motions~\cite{wang2024continuous}, limiting current GT to 120 Hz. To break this bottleneck without relying on complex commercial rigs or labor-intensive annotations, we propose FlashCap. Our portable LED design provides an automated, studio-free approach for bandwidth-friendly 1000Hz GT acquisition of high-speed micro-dynamics (see Appendix Tab. 19).

We develop FlashCap, the first flashing LED-based MoCap system to capture human motion at high temporal resolution with low bandwidth and monetary cost. It includes a motion capture suit with multiple flashing LEDs and IMUs, as well as a multimodal capture device comprising an RGB camera and an event camera. We develop a label-annotation process to generate high-frequency ground-truth labels from the flashing light emitted by LEDs. 

Based on FlashCap, we collect FlashMotion, a dataset with high-temporal-resolution labels. 
\emph{The dataset is recorded with 1000~Hz ground truth labels, improving upon the state-of-the-art temporal resolution (120\; Hz) in public HPE datasets by nearly an order of magnitude.} We evaluate the dataset's quality by comparing it against motions recorded by a high-speed RGB camera and manual annotations.

To evaluate the merits of FlashMotion, we propose two novel HPE tasks: precise motion timing (PMT) and high-temporal-resolution HPE. PMT estimates the time of a specific motion, while high-temporal-resolution HPE performs HPE at millisecond resolution. These two tasks are critical for swift sport analysis.

Bridging the gap between standard low-frame-rate inputs and the proposed 1000\, Hz ground truth poses a significant challenge for existing frame-based methods. To address this, we develop \textbf{ResPose}, a simple yet effective baseline. It treats high-frequency events as residual signals to refine static RGB anchors. ResPose leverages residual poses to demonstrate a viable path toward millisecond-level tracking — a precision level beyond the temporal resolution of existing public benchmarks.

The experimental results show that while FlashMotion brings significant challenges to the community, our proposed baseline establishes a strong initial benchmark, reducing the Mean Per Joint Position Error (MPJPE) by $\sim$40\% compared to standard RGB interpolation. We hope that FlashMotion could provide opportunities for a deep understanding of human motion at high temporal resolution.

\section{Related Work}
\label{sec:formatting}


%

\subsection{Motion Capture Techniques for Pose Labels}
Ground-truth labels play a pivotal role in HPE research. Current methodologies for acquiring labels primarily fall into two categories: RGB-based and inertial measurement unit (IMU)-based solutions. Multi-view methods, such as Vicon~\cite{VICON}, use multi-camera studios with retro-reflective markers to obtain labels. They are expensive and have limited acquisition sites. Monocular RGB approaches~\cite{VIBE, SIMPX} employ neural networks at the expense of reduced precision. A fundamental limitation of optical methods is their temporal resolution, which is constrained by the frame rate of RGB cameras. Conventional cameras operate at 30-60Hz. Although high-speed variants ($\geq 1000$ Hz) exist, they require extremely strong lighting conditions to maintain good acquisition quality~\cite{EventCap}. The bandwidth requirements of high-speed cameras can be two orders of magnitude higher than those of event cameras~\cite{natureEvent}, placing significant computational and storage burdens. Moreover, high-speed RGB cameras are costly. For example, a high-speed RGB camera (i.e., NAC Memrecam HX-7s) costs above 45,000 USD, which is about 9 times that of a Prophesee event camera.

IMU (Inertial Measurement Unit)-based alternatives, exemplified by Noitom~\cite{Noitom} and Xsens~\cite{XSENS}, circumvent optical constraints but introduce drifting errors~\cite{HSC4D}. Moreover, their frequency is lower at 1000 Hz. 

\subsection{HPE Method and Datasets}

Researchers have adopted RGB-based~\cite{PROX, cai2022humman, shao2022diffustereo,cheng2023dna,HUMOR_ICCV2021, MotionBert2023,PARE_ICCV2021}, IMU-based~\cite{pons2011outdoor,Yi2021TransPoseR3,SIP,DIP}, LiDAR-based~\cite{lidarcap,HSC4D}, WIFI-based~\cite{WiPoseMobiCom20,WinectRen2021,MetaFiIOJ2023}, Radar-based~\cite{an2022mri,chen2022mmbody,chen2023immfusion}, electromagnetic-based~\cite{kaufmann2023emdb}, Event-based~\cite{EventCap,lifting2021,TORE2023}, or multimodal-based~\cite{chen2021sportscap,Bogo2016KeepIS, MAED, wang2023nemo, GFPose2023,Su2020RobustFusionHV,yang2021s3,Patel2020TailorNetPC, sun2023trace,relid11d,ascendmotion} methods to estimate human poses. We compare our work with multiple datasets in Tab.~\ref{tab:data_compare}. To our knowledge, FlashMotion is the \textbf{first publicly available multi-modal human motion dataset with millisecond-accuracy pose labels}, 
boasting the highest annotation framerate (1000 FPS)—an order of magnitude beyond state-of-the-art datasets, the largest labeled frame count (7150k frames), enabling unprecedented high-temporal-resolution analysis, and a large 3D labeled frame count, making it one of only two large-scale datasets (>=1 hour) with high-frequency 3D labels (>=60 FPS).

\begin{table*}[!t]
    \caption{Comparisons with related human motion datasets. Datasets with Event camera data are grouped in the second block. Our FlashMotion provides the highest 2D label frame rate (1000Hz) and the largest number of labeled frames (7.15M) by a significant margin. Abbreviations: (Dur.) Duration, (\# Seqs.) Number of Sequences, (\# Subjs.) Number of Subjects.}
    \vspace{-1mm}
    \label{tab:data_compare}
    \centering
    \resizebox{1\linewidth}{!}{
    \begin{tabular}{l cccc cc c ccc}
        \toprule[1pt]
        
        \multirow{2}{*}{\textbf{Dataset}} & \multicolumn{4}{c}{\textbf{Sensor Modalities}} & \multicolumn{2}{c}{\textbf{Label FPS}} & \multirow{2}{*}{\textbf{Dur.}} & \multirow{2}{*}{\textbf{Label Frames (k)}} & \multirow{2}{*}{\textbf{\# Seqs.}} & \multirow{2}{*}{\textbf{\# Subjs.}} \\
        \cmidrule(r){2-5} \cmidrule(r){6-7}
        & RGB & MoCap & LiDAR & Event & 2D & 3D & & & & \\
        
        \midrule
        
        LiDARHuman26M~\cite{lidarcap} & \CheckmarkBold & IMU & \CheckmarkBold & - & - & 10 & 5h & 184 & 20 & 13 \\
        HSC4D~\cite{HSC4D} & - & IMU & \CheckmarkBold & - & - & 20 & 42min & 50 & 8 & 1 \\
        SLOPER4D~\cite{dai2023sloper4d} & \CheckmarkBold & IMU & \CheckmarkBold & - & 30 & 20 & 2.8h & 300 & 15 & 12 \\
        CIMI4D~\cite{yan2023cimi4d} & \CheckmarkBold & IMU & \CheckmarkBold & - & - & 20 & 2h & 180 & 42 & 12 \\
        LIPD~\cite{ren2023lidar} & \CheckmarkBold & IMU & \CheckmarkBold & - & - & 10 & 1.7h & 62 & 10 & 6 \\
        EMDB~\cite{kaufmann2023emdb} & \CheckmarkBold & EM & - & - & - & 30 & 58min & 105 & 81 & 10 \\
        FreeMotion~\cite{freemotion} & \CheckmarkBold & IMU & \CheckmarkBold & - & 60 & 60 & - & 578 & - & 20 \\
        AscendMotion~\cite{ascendmotion} & \CheckmarkBold & IMU & \CheckmarkBold & - & 20 & 20 & 13h & 412 & 220 & 22 \\
        BEDLAM~\cite{black2023bedlam} & \CheckmarkBold & - & - & - & 30 & 30 & - & 1000 & - & - \\
        
        \midrule
        
        DHP19~\cite{calabrese2019dhp19} & - & - & - & \CheckmarkBold & 100 & 100 & 3.3h & 1178 & 33 & 17 \\
        EventCap~\cite{EventCap} & \CheckmarkBold & - & - & \CheckmarkBold & - & 100 & 58s & 1.42 & 12 & 6 \\
        MMHPSD~\cite{EventHPEICCV21} & \CheckmarkBold & - & - & \CheckmarkBold & 15 & 15 & 4.5h & 234 & 84 & 15 \\
        EHPT-XC~\cite{NeurIPS24FengGuanJin} & \CheckmarkBold & - & - & \CheckmarkBold & 30 & - & 13min & 16 & 158 & 82 \\
        CDEHP~\cite{shao2024temporal} & \CheckmarkBold & - & - & \CheckmarkBold & 60 & 60 & 23min & 82 & 25 & 20 \\
        BEAHM~\cite{wang2024continuous} & \CheckmarkBold & - & - & \CheckmarkBold & 120 & 120 & - & - & 40 & - \\
        RELI11D~\cite{relid11d} & \CheckmarkBold & IMU & \CheckmarkBold & \CheckmarkBold & 20 & 20 & 3.3h & 239 & 48 & 10 \\
        
        \midrule

        \textcolor[rgb]{0.30,0.53,1.00}{\textbf{FlashMotion(Ours)}} & 
        \textcolor[rgb]{0.30,0.53,1.00}{\CheckmarkBold} & 
        \textcolor[rgb]{0.30,0.53,1.00}{\textbf{LED}} & 
        \textcolor[rgb]{0.30,0.53,1.00}{\CheckmarkBold} & 
        \textcolor[rgb]{0.30,0.53,1.00}{\CheckmarkBold} & 
        \textcolor[rgb]{0.30,0.53,1.00}{\textbf{1000}} & 
        \textcolor[rgb]{0.30,0.53,1.00}{\textbf{60}} & 
        \textcolor[rgb]{0.30,0.53,1.00}{\textbf{2h}} & 
        \textcolor[rgb]{0.30,0.53,1.00}{\textbf{7150}} & 
        \textcolor[rgb]{0.30,0.53,1.00}{\textbf{240}} & 
        \textcolor[rgb]{0.30,0.53,1.00}{\textbf{20}} \\
        
        \bottomrule[1pt]
    \end{tabular}%
    }
\end{table*}

 
 Event cameras offer high frame rates at low cost, with minimal storage and bandwidth. These features enable event cameras to effectively capture high-speed human motions. 
 Multiple event-based HPE are developed, such as EventCap~\cite{EventCap}, EventPointPose~\cite{EventPointPose3DV22}, EventHPE~\cite{EventHPEICCV21}, and \cite{aydin2024hybrid}.

 DHP19~\cite{calabrese2019dhp19} provides 100 Hz Vicon ground-truth poses for multi-view event-based human pose estimation.
 EventCap~\cite{EventCap} collects human pose with 100 FPS ground truth labels. 
 MMHPSD~\cite{EventHPEICCV21} uses five RGB-D cameras recording at 15 FPS. LiftMono-HPE\cite{lifting2021} collects event-based motion data through simulation. 
 EHPT-XC~\cite{NeurIPS24FengGuanJin} captures low-light high-speed motion via event and dual RGB cameras, with 20Hz pose labels from RGB.
 CDEHP~\cite{shao2024temporal} is an event-based dataset under changing lighting conditions, with RGB-derived pose labels limited to 60Hz. 
 BEAHM~\cite{wang2024continuous} is an event-based pose dataset employing multi-view reconstruction via four calibrated 120Hz RGB cameras. 
While event-based datasets like DHP19~\cite{calabrese2019dhp19} (100\, Hz Vicon), EventCap~\cite{EventCap} (100\, Hz markerless), and BEAHM~\cite{wang2024continuous} (120\, Hz RGB) have advanced the field, their ground truth relies on traditional optical systems, inherently bottlenecked by camera frame rates.

FlashCap introduces a paradigm shift: by encoding identity directly into high-frequency LED blinking patterns, our system generates native 1000\, Hz pose labels directly from the event stream. This bypasses the frequency limitations of conventional MoCap, enabling the first millisecond-level benchmark.
HHMotion~\cite{HHMotion} also highlights the need for fine-grained motion evaluation in robotics.

\section{FlashCap: flashing LED-based MoCap}
\label{section: FlashCap}



There is a lack of high-temporal-resolution (i.e., 1000Hz) multi-modal human motion datasets, which limits HPE applications such as precise motion timing. To capture human motions with high-temporal-resolution ground truth labels, we build FlashCap, a flashing LED-based MoCap system. It includes a motion-capture outfit and a multi-modal capture device. We also develop a data annotation pipeline to annotate poses with high temporal resolution.

\begin{figure}[!tb]
    \centering
     \includegraphics[width=1\linewidth]{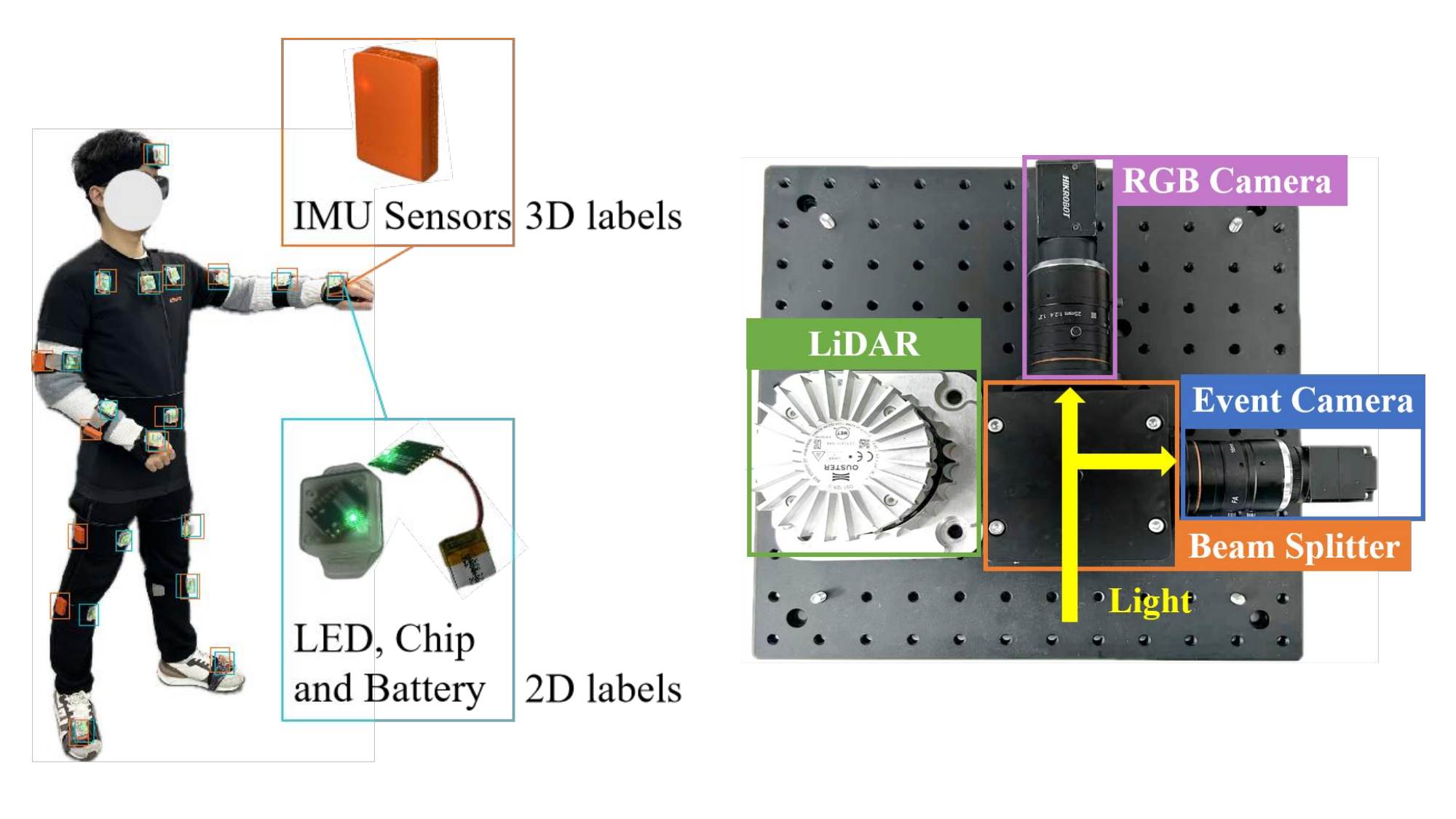}
     \vspace{-5mm}
     \caption{The FlashCap Mocap outfit(left) and multi-modal capture device(right).}
     \label{fig:suit_and_device}

\end{figure}

\subsection{MoCap Outfit}


Fig.~\ref{fig:suit_and_device} depicts the FlashCap motion capture suit. The MoCap suit contains 17 LEDs and 17 IMUs. These IMUs come from an Xsens MVN inertial motion capture system. 

The LEDs are attached to different parts of the human body. Specifically, each LED is mounted on a small chip that controls it to emit green light. A small battery powers each chip. An LED, chip, and battery are installed in a 3D-printed small transparent plastic box. The plastic box is attached to a strap (such as a limb) or clothing (such as a shoulder) with Velcro. Since each LED is firmly attached to a distinct body part, we use the LED pixel locations as the body part locations when generating high-temporal-resolution labels.
We place markers on the relatively stable mid-segments of bones to avoid significant displacement caused by joint articulation, thereby ensuring consistent tracking targets.



Each LED \(i\) emits a flash of light at a configurable frequency (e.g., 4000 Hz) and serves as a marker, easily detected by an event camera as a stream of events. When a pixel's illumination exceeds a predefined threshold, the event camera asynchronously triggers an event. An event is represented as $e = (h, w, t, p)$, where \(h\) and \(w\) denote the pixel locations, \(t\) is the event timestamp, and \(p\) is the polarity. 

In each flicker period $T_i$ of LED $i$, the LED is on for $t^p_i$ duration, and off for $t_i^n$ duration. We refer to $t^p_i$ and $t^n_i$ as the on-time and the off-time for LED $i$, respectively. To ease the identification of different LEDs, we configure the on-time and off-time of them differently, ranging from 100\,\(\mu\)s and 300\,\(\mu\)s. \emph{For details on LED color selection, placement, and configuration, please refer to the Supplementary.}




\subsection{A Multi-modal Motion Capture Device}

The FlashCap multi-modal motion capture device is depicted in Fig.~\ref{fig:suit_and_device}. The multi-modal capture device consists of an RGB camera (Hikrobot $1920\times1200$) and an event camera (Prophesee $1280\times720$). Additionally, we use a LiDAR (Ouster OS-1 128-beam) to record 3D point clouds at 20 FPS. The RGB camera captures RGB frames at 20 FPS. 


Our system captures high-speed motion data using event cameras and LEDs. While LiDAR and IMU are integrated for verification, they are not essential for operation and can be removed in deployment to lower costs.

\textbf{Time Synchronization and Calibration.} To ensure precise multi-modal alignment, we perform rigorous temporal synchronization and spatial calibration across all sensors, including employing a beam splitter for event-RGB pixel alignment. Comprehensive details are provided in the Supplementary Material.

\subsection{Data Annotation Pipeline}\label{sec:flashcap:annotation}
\begin{figure*}[tb]
    \centering
    \includegraphics[width=1\linewidth]{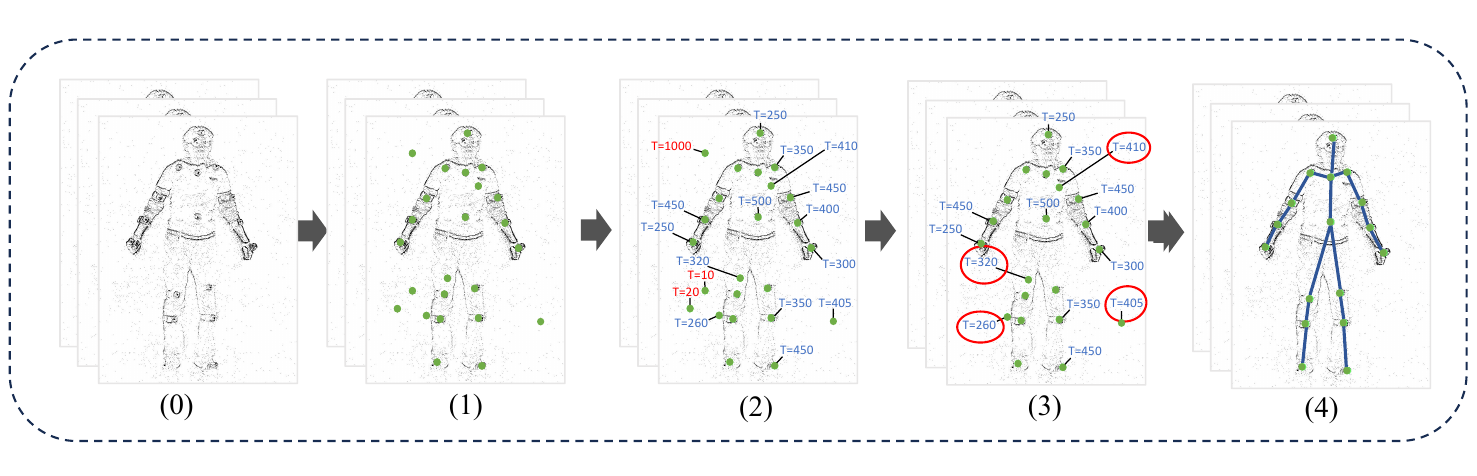}
    \vspace{-8mm}
    \caption{An example of the FlashCap data annotation pipeline: (0) Event Streams. (1) Identified Event Clusters. (2) Cluster Frequency Analysis. (3) Filtered Clusters After Noise Removal. (4) Matched LED-Cluster Pairs (Labels).} 
    \label{fig:pipeline}
 \end{figure*}



The flashing light emitted by LED $i$ forms an event cluster $j$, which contains events that are close spatially and temporally. The annotation pipeline seeks to identify the correct mapping between LED $i$ and cluster $j$. 

Given a set of identified event clusters $[c_j]_{j=1}^M$, we want to find its corresponding LEDs $[LED_i]_{i=1}^N$, where $M$ and $N$ are the number of clusters and LEDs, respectively. Once the correct mapping is found, the location of a human joint is identified as the centroid of the matched event cluster.  

An example of the data annotation pipeline is depicted in Fig.~\ref{fig:pipeline}. It consists of four steps: (1) event cluster identification (Sec .~\ref {sec:flashcap:annotaion:cluster}), (2) cluster frequency identification (Sec .~\ref {sec:flashcap:annotaion:clusterFreq}), (3) outlier filtering and noise removal (Sec .~\ref {sec:flashcap:annotaion:outlier}), and (4) LED and cluster matching (Sec .~\ref {sec:flashcap:annotaion:matching}).
To ensure the quality of the annotation labels, we develop an annotation tool for human annotators to improve the quality of labels generated by previous steps. Please refer to the Supplementary for details of the tool.


\subsubsection{Event Cluster Identification} \label{sec:flashcap:annotaion:cluster}

The event streams are partitioned into event frames over fixed time intervals (e.g., 1 millisecond), transforming the asynchronous event stream into a series of spatiotemporal snapshots. These frames are then processed using a density-based clustering algorithm (i.e., DBSCAN\cite{ester1996density}) to form clusters corresponding to a potential LED joint location, as the flashing LEDs generate high-density event regions. The distance threshold is set to ensure a 99\% recall rate, ensuring that almost all joints are detected. This high recall design ensures minimal loss of accurate positive detections, even at the cost of introducing a few redundant or overlapping clusters, which will be filtered in subsequent steps. We have studied multiple clustering methods and different hyperparameters. Please refer to the Supplementary for details.

\subsubsection{Cluster Frequency Identification}\label{sec:flashcap:annotaion:clusterFreq}

In each event frame, we count the number of positive polarity events and negative polarity events. If the number of positive polarity events exceeds that of negative polarity events, the event frame is classified as positive; otherwise, negative. This process yields a complete event sequence \(E_{seq}\), which consists of positive and negative event windows ordered by time.
By analyzing polarity changes in \(E_{seq}\), we compute the average on-time (\(\overline{t^p_j}\)) and average off-time (\(\overline{t^n_j}\)), as well as the flicker period \(\overline{T^j} = \overline{t^p_j} + \overline{t^n_j}\) for cluster $j$ by averaging the time differences (\(\Delta t\)) between successive polarity changes.

\subsubsection{Outlier Filtering and Noise Removal}\label{sec:flashcap:annotaion:outlier}

Due to environmental disturbances and sensor limitations, noisy events may lead to the incorrect identification of event clusters. To ensure joint identification accuracy, we apply temporal smoothing to mitigate abrupt changes in polarity sequences induced by transient environmental interference. Further, we apply outlier filtering to remove noise-induced fluctuations when estimating \(t_p^j\), \(t_n^j\), and \(T_j\). An outlier cluster is defined as a cluster whose time values (e.g., $t^p_j$, $t^n_j$, and $T_j$) deviate significantly from the expected LED blinking pattern.


    
    




        


\subsubsection{LED and Cluster Matching} \label{sec:flashcap:annotaion:matching}
For the clusters $[c_j]_{j=1}^M$, we use a bipartie algorithm to find its corresponding LEDs $[LED_i]_{i=1}^N$. The algorithm requires the distance between clusters and LEDs. The distance $d_{ji}$ among event cluster $j$ and LED $i$ is calculated as follows.

\begin{enumerate}
    \item \textbf{On-off time distance} $    d_{ji}^t = |\overline{t^p_j} - t^p_i| + |\overline{t^n_j} - t^n_i|$,    where \(\overline{t^p_j}\) denotes the average on-time of cluster $j$, \(t^p_i\) the on-time of LED $i$, \(\overline{t^n_j}\) the average off-time of cluster $j$, and \(t^n_i\) the off-time of LED $i$.


  \item \textbf{Period distance} $    d^p_{ji} = |T^n_j - T_i| + |T^p_j - T_i|$,    where \(T^n_j\) and \(T^p_j\) represent the periods of the negative and positive polarity event sequences, respectively, and \(T_i\) is the flicker period of LED $i$.

    \item \textbf{Cluster and LED distance} $d_{ji} = \alpha \cdot d^t_{ji} + \beta \cdot d^p_{ji}$,     where \(\alpha\) and \(\beta\) are weighting coefficients.

\end{enumerate}

\textbf{Tracking} Some event clusters may be wrongly removed due to noise. To ensure stable matching, we devise a tracking algorithm to track a set of identified event clusters from previous frames. 




We have demonstrated the robustness of FlashCap across lighting conditions and LED occlusions, and we have discussed the impact of noise on events in the Supplementary. 
\section{The FlashMotion Dataset}
\subsection{Dataset Description}

\begin{figure}[!tb]
    \centering
     \includegraphics[width=1\linewidth]{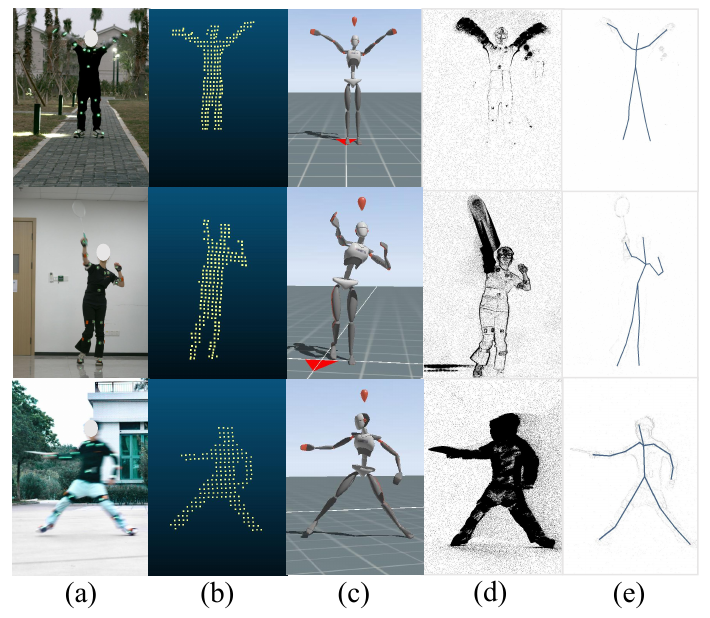}
     \vspace{-5mm}
     \caption{Examples of the FlashMotion dataset. (a) RGB modality, (b) LiDAR point clouds, (c) IMUs measurements, (d) events (50ms), (e) 2D Label/Annotation. }
     \label{fig:dataset}
\end{figure}




Based on the FlashCap motion capture system, we collect a human motion dataset, FlashMotion, with high-temporal-resolution labels. The FlashMotion dataset consists of 240 sequences in 4 modalities, including 144,350 RGB frames, 144,350 LiDAR point cloud frames, and 2 hours of IMU data and events.  It contains 2D labels recorded at 1000Hz.  To our knowledge, this is the first dataset that contains such high-temporal-resolution human pose ground truth without using a high-speed RGB camera. 
To collect FlashMotion, we invite 20 volunteers (10 females and 10 males) to perform swift motions in 4 scenes, indoor and outdoor. The dataset includes 11 major action categories and 19 detailed action types. Please refer to the Supplementary for details. Examples of the FlashMotion dataset are depicted in Fig.~\ref{fig:dataset}. 
\begin{figure}[t]
    \centering
    \includegraphics[width=1\linewidth]{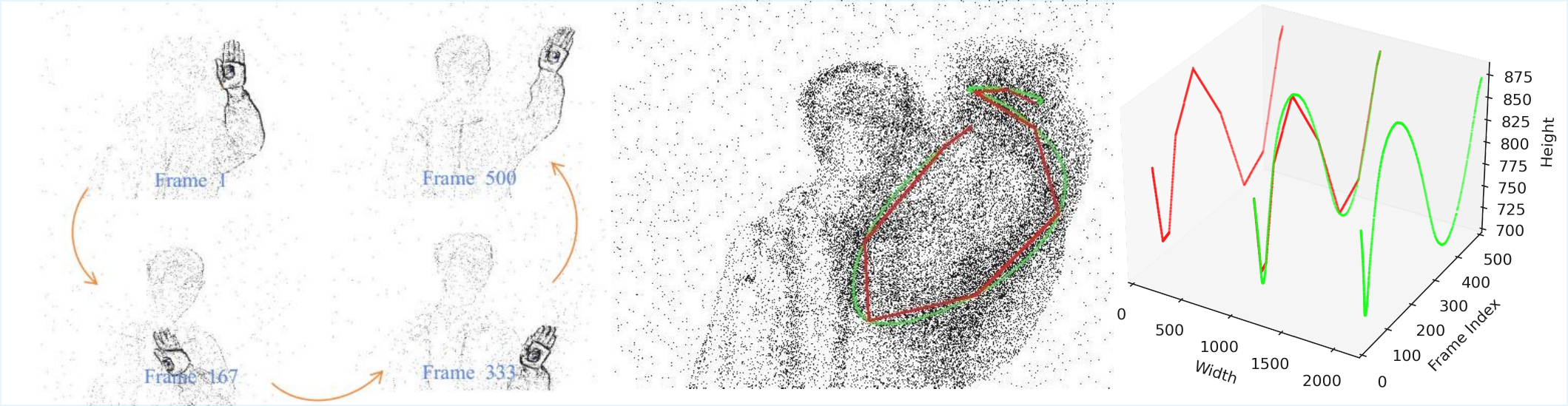}
    \vspace{-4mm}
    \caption{Interpolated vs. Ground-Truth 1000\,Hz Poses. (Left) Event frames at selected timestamps. (Middle) Accumulated event stream. (Right) trajectory comparison. The red line indicates 20\,Hz interpolation, and the green line indicates 1000\,Hz ground truth.}
    \label{fig:wave}
\end{figure}

There are two types of motion labels in the FlashMotion dataset: the 60Hz 3D human pose labels and 1000Hz 2D human pose labels. For 3D labels, we employ the approach of~\cite{relid11d} to process the IMU data and LiDAR data to obtain 60 Hz SMPL~\cite{SMPL2015} parameters. The 3D labels are used for verification purpose, and they are shared with the research community for a better understanding of human motions. For 2D labels, we use the data annotation pipeline in Sec.~\ref{sec:flashcap:annotation} to obtain the initial 1000 Hz 2D joint point labels and then use the human annotation tool (see Supplementary) for manual correction to fix a small percentage of erroneous labels. 
Beyond elite sports~\cite{Luge}, millisecond precision is vital for general rapid motions. Fig.~\ref{fig:wave} shows that 20,Hz interpolation fails to match our 1000,Hz ground truth during rapid hand-waving. Furthermore, even spline interpolation from 100 Hz RGB cameras yields significant Max Errors (ME) in highly dynamic actions (\textit{Jumping}: \textbf{28.5,px}; \textit{Rotating Hands}: \textbf{11.2,px}). This confirms standard high-speed cameras miss micro-dynamics, validating the necessity of our 1000 Hz event-based ground truth (see Appendix~A).
\subsection{Dataset Evaluations}
\begin{figure}[!tb]
    \centering
     \includegraphics[width=1\linewidth]{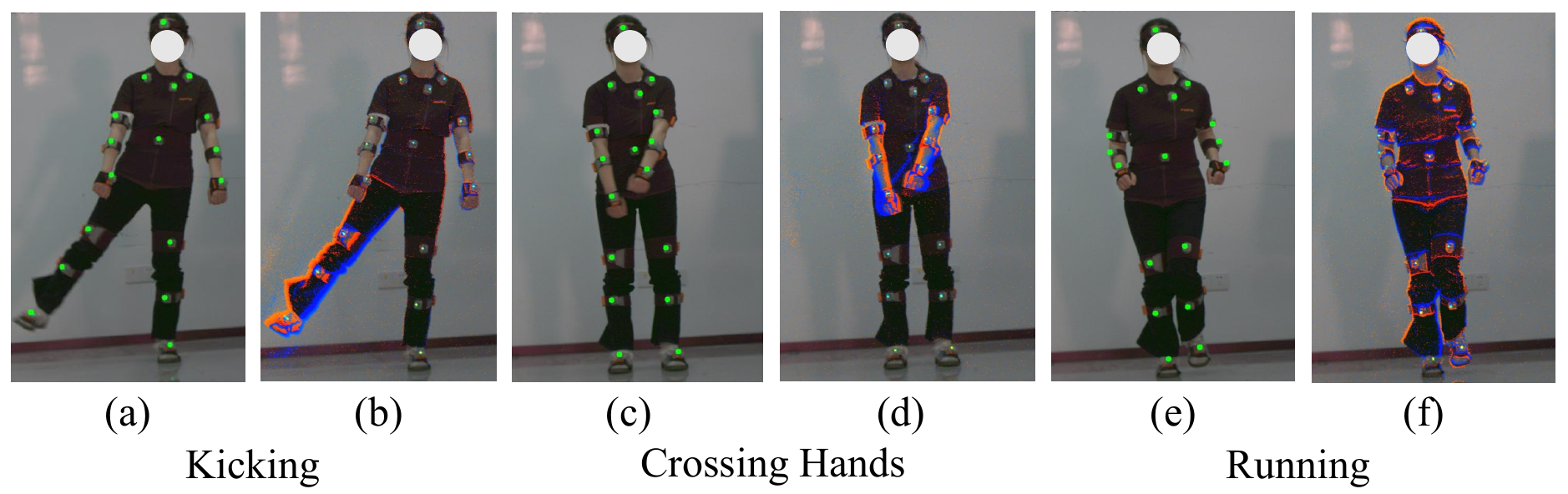}
     \vspace{-4mm}
     \caption{
     Qualitative evaluation against a high-speed camera: Columns (a), (c), and (e) show that our labels (green spots) overlap with the images captured by high-speed cameras closely. This demonstrate the correctness of FlashMotion labels qualitatively. Columns (b), (d), and (f) show that the corresponding events overlaid on the RGB frames. The red/blue dots indicate events with positive/negative polarity.}
     \label{fig:qualitative_eval}
\end{figure}

\begin{figure}[!tb]
    \centering
     \includegraphics[width=1\linewidth]{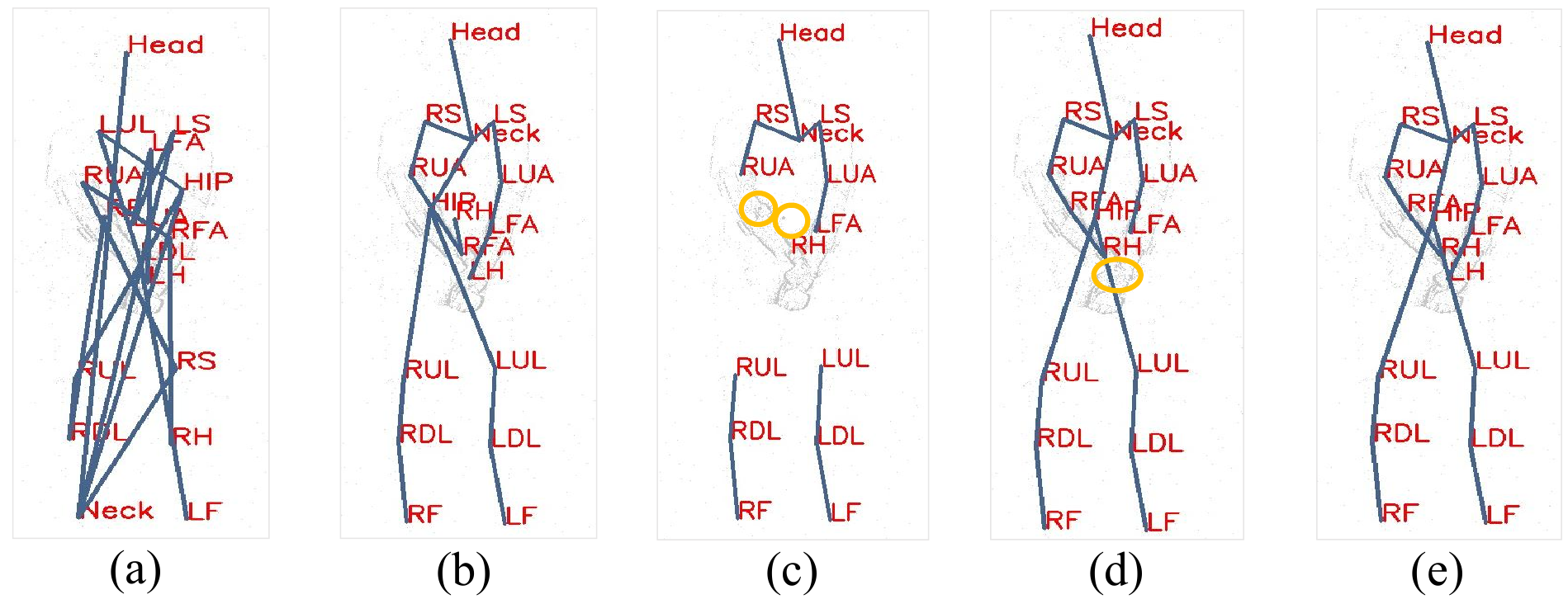}
     \vspace{-4mm}
     \caption{Qualitative evaluation when ablating the annotation pipeline: (a) $w/o$ $d_{ji}^t$ in Sec.~\ref{sec:flashcap:annotaion:matching}  (b) $w/o$ $d^p_{ji}$ in Sec.~\ref{sec:flashcap:annotaion:matching}. (c) $w/o$ Outlier Filtering (Sec.~\ref{sec:flashcap:annotaion:outlier}). (d) $w/o$ Tracking. (e) complete pipeline. The name of the identified joints are depicted in red near the location of identified joints. Each joints are connected by blue lines, and the yellow circles in (c) and (d) indicates missed labels due to failed detection and occlusion, respectively.}
     \label{fig:qualitative_eval_pipeline}
\end{figure}


In this section, the quality of the FlashMotion dataset is demonstrated. For evaluation, a volunteer performs Kicking, Crossing Hands, and Running motions. 
To evaluate FlashMotion's quality, a volunteer performs Kicking, Crossing Hands, and Running. We first qualitatively validate the 2D labels and events against high-speed RGB captures. Then, we qualitatively and quantitatively ablate our annotation pipeline against manual labels.

For evaluation, we replace the standard RGB camera with a Basler high-speed RGB camera, which captures high-frame-rate (100 Hz) RGB video with $2448\times2048$ resolution. The high-speed RGB camera and the event camera share the same view via the beam splitter and are time-synchronized. The labels generated by the automatic labeling pipeline and the events are superimposed on the high-speed frames. The timestamps of the labels and events match those of the RGB frames recorded by the high-speed RGB camera. The qualitative evaluation is shown in Fig.~\ref{fig:qualitative_eval}. 

We analyze annotation-pipeline components in Fig.~\ref{fig:qualitative_eval_pipeline}. Removing matching terms ($d_{ji}^t$ or $d^p_{ji}$) causes joint misclassification (a-b), while removing outlier filtering or tracking leads to missed joints (c-d). We evaluate 24 manually annotated sequences across 8 action categories (10\% of the dataset). A prediction is counted as correct if within 1 pixel of human annotation. The complete pipeline achieves 99.99\% precision and 98.82\% recall (Tab.~\ref{tab:ablation:human}), demonstrating strong robustness; per-category results are in the Supplementary.

\begin{figure*}[tb]
    \centering
    \includegraphics[width=1\linewidth]{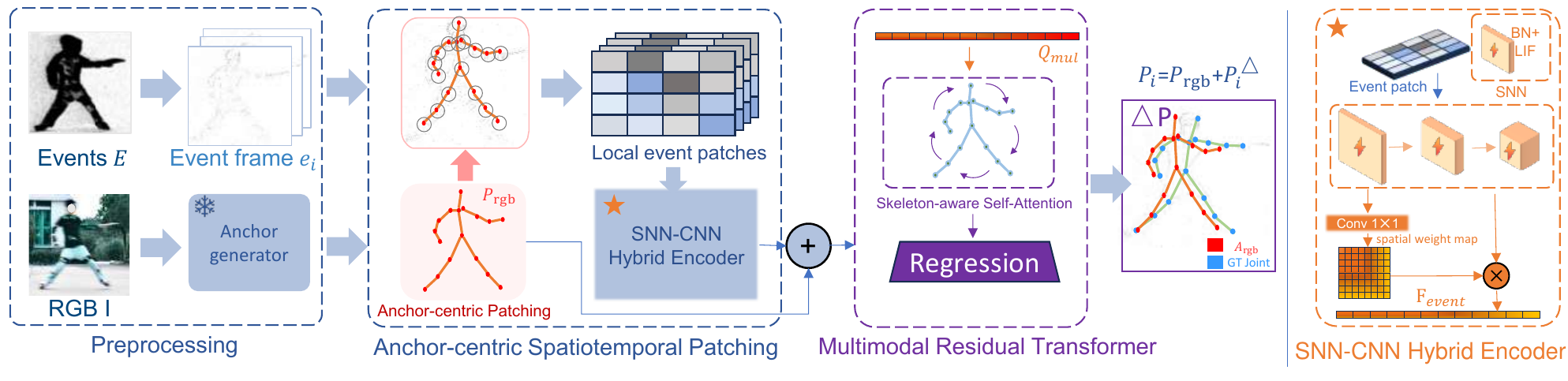}
    \vspace{-5mm}
\caption{\textbf{Architecture of ResPose.} It obtain high-temporal resolution pose $P_i$ through $P_i = P_{\text{rgb}} + P_i^{\Delta}$, where $P_{\text{rgb}}$ is the anchor pose from the RGB branch and $P_i^{\Delta}$ is the residual pose at time $i$}%

    \label{fig:baseline}
 \end{figure*}

\section{ResPose: A High-temporal Resolution HPE}
\label{sec:baseline}

In this work, we develop ResPose, a simple, high-temporal-resolution pose estimation method. ResPose consists of two input branches: a low-frame-rate RGB branch and a high-frame-rate event branch. It leverages the structural stability of RGB priors while exploiting the high temporal resolution of events to capture micro-movements.

At time interval (e.g. every 50 milliseconds), given an RGB frame $I$ and a corresponding stream of asynchronous events $[e_i]_{i=0}^N$, ResPose estimates a 2D major pose $P_{\text{rgb}}$ base on RGB frame $I$ through an RGB HPE method (e.g., ViTPose~\cite{vitpose}) and estimates a series of residual pose $[P_i^{\Delta}]_{i=0}^N$. $N$ is the number of event frames in the time interval. ResPose obtains the high-temporal resolution pose $[P_i]_{i=0}^N$ through $P_i = P_{\text{rgb}} + P^{\Delta}_i$.

\textbf{SNN-CNN Hybrid Encoder} The high-frequency event branch is tasked with extracting micro-movement dynamics. Driven by the inductive bias that micro-movements occur near the joints, the encoder performs dynamic cropping centered at each normalized RGB anchor $P_{\text{rgb}}$. This operation extracts local spatiotemporal event patches ($32 \times 32$) at each micro-time step, explicitly filtering out task-irrelevant background disturbances.

The encoder employs \textit{Leaky Integrate-and-Fire (LIF)} neurons for temporal spike integration. To mitigate background noise, it uses a lightweight $1 \times 1$ convolutional branch which highlights informative motion regions while attenuating task-irrelevant clutter. Prior work has demonstrated that SNNs are highly efficient for processing the asynchronous event modality~\cite{HsVT,SpikeSlicer}. Feature maps are unrolled directly into $F_{\text{event}} \in \mathbb{R}^{J \times 4096}$.

\textbf{Multimodal Residual Transformer} To enforce kinematic consistency and fuse heterogeneous modalities, we employ a Transformer encoder as the global residual regressor. The 2D anchors $P_{\text{rgb}}$ are lifted via a learnable linear layer and concatenated with $F_{\text{event}}$,  forming a unified embedding that encapsulates both RGB structural priors and event motion tendencies.


\textbf{Skeleton-aware Self-Attention.} The Transformer models global dependencies among the $J=17$ joints, serving as a kinematic constraint to rectify noisy local estimates. The entire baseline is optimized end-to-end by minimizing the $L_2$ distance between ground truth and predicted poses.



\section{Tasks and Benchmarks}
To evaluate the merit of FlashMotion, we conduct rigorous benchmarking against two novel HPE tasks: Precise motion timing and high temporal resolution HPE. Please refer to the Supplementary for detailed setup and metrics.

\begin{table}[!tb]
    \caption{Quantitative evaluation against human annotated labels.}
    \centering
    \scriptsize
    \resizebox{\linewidth}{!}{
    \begin{tabular}{c|cc|cc}
        \toprule
        \multirow{2}{*}{Method} 
        & \multicolumn{2}{c|}{Kicking} 
        & \multicolumn{2}{c}{Crossing Hands}  \\
        & Precision$\uparrow$ & Recall$\uparrow$ 
        & Precision$\uparrow$ & Recall$\uparrow$ \\
        \midrule
        $w/o$ \(d_{ji}^t\) & 43.34\% & 97.80\% & 23.64\% & 98.54\%  \\
        $w/o$ \(d^p_{ji}\) & 69.70\% & 97.56\% & 64.00\% & 98.33\% \\
        $w/o$ Outlier Filtering & 96.52\% & 95.69\% & 83.17\% & 98.50\%  \\
        $w/o$ Tracking & 98.38\% & 98.16\% & 84.29\% & 98.21\%  \\
        Complete pipeline & 99.99\% & 98.99\% & 99.99\% & 98.65\%  \\
        \bottomrule
    \end{tabular}
    }
    \label{tab:ablation:human}
\end{table}


    

    

    
\begin{table}[!tb]
    \caption{The Mean Error of Estimated Time (PMT).Unit: $ms$.}
    \label{tab:timing}
    \centering
    \footnotesize{
    \begin{tabular}{l ccc}
        \toprule
        Method & Kicking & Punching & Jumping \\
        \midrule 
        \centering ViTPose~\cite{vitpose} & 48.5 & 62.3 & 31.4 \\
        
        \centering Hybrid ANN-SNN~\cite{aydin2024hybrid} & 85.2 & 54.1 & 66.7 \\

        \centering LEIR~\cite{relid11d} & 112.4 & 135.8 & 78.2 \\
        \midrule
        
        \centering \textbf{ResPose (Ours)} & \textbf{7.2} & \textbf{4.8} & \textbf{6.5} \\
        \bottomrule
    \end{tabular}%
    }
\end{table}
\subsection{Precise Motion Timing (PMT)}

Accurate timing of human motion is important for swift motion analysis, such as punching and jumping speed calculation. Given a sequence of motions, precise timing tasks measure the time when a specific joint passes a line, which can be used to measure reaction time and speed. 

\begin{figure}[!tb]
    \centering
     \includegraphics[width=1\linewidth]{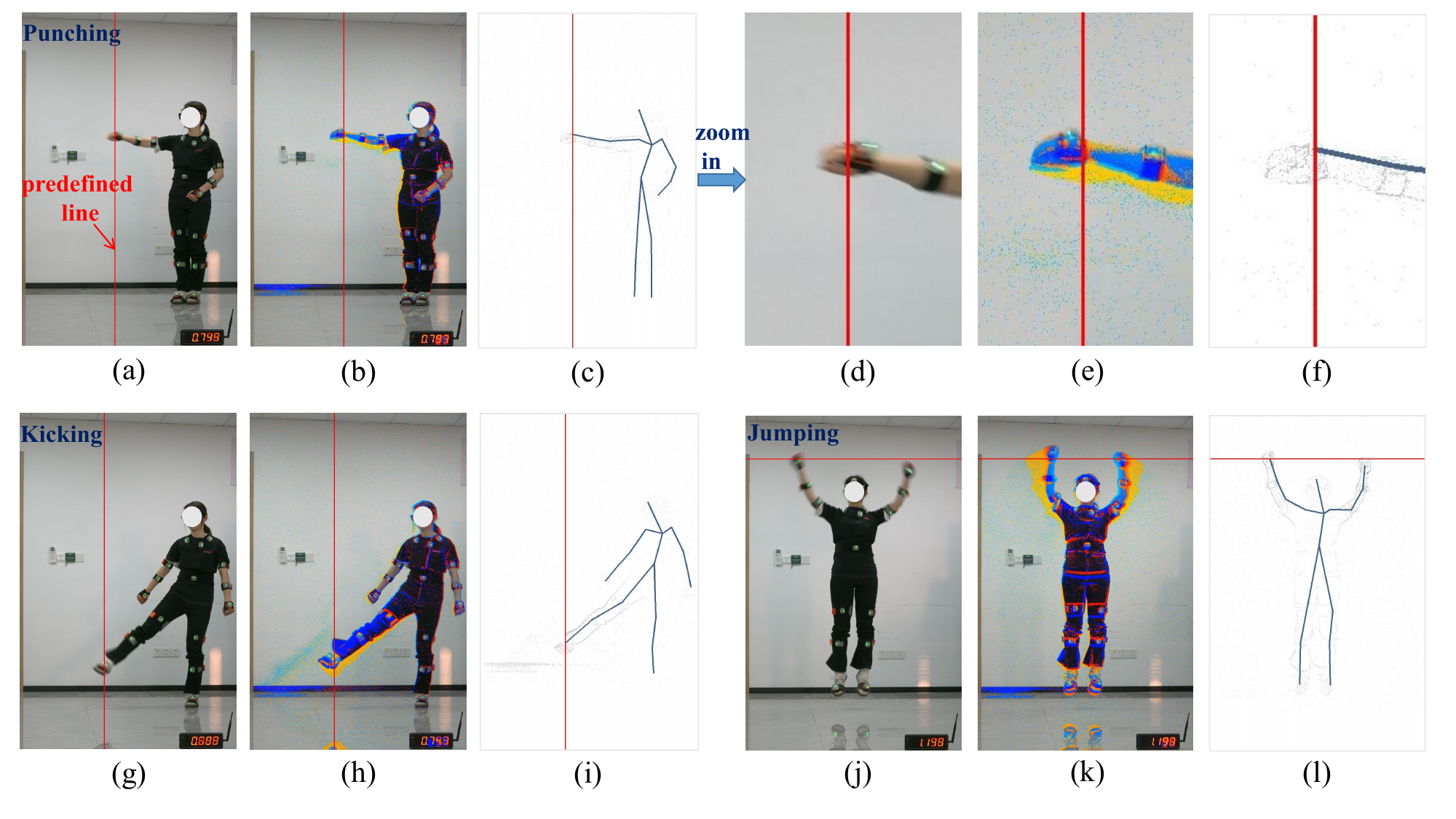}
     \vspace{-5mm}
     \caption{Precise timing. (a) (d) (g) (j): Previous RGB frame just before crossing lines, (b) (e) (h) (i): RGB frame of corresponding events just after crossing the lines, (c) (f) (i) (l): The event and human pose at the moment when crossing the line. (a)-(f): Punching. (g)-(i): Kicking. (j)-(l): Jumping.}
     \label{fig:timing}
\end{figure}

For such tasks, volunteers perform motions when receiving a visual signal (flag waving). We place a millisecond meter in front of our multi-modal acquisition device. The millisecond meter starts timing when it detects the flag-waving signal for performing motions, and we use the moment as the starting moment of the movement. The volunteer performs different actions to make her/his pre-specified joint pass through a pre-defined line. When the joint passes the line, the moment is considered the end of the movement. Some inputs of this task are shown in Fig.~\ref{fig:timing}.For fairness, all methods use the same start signal, the same predefined crossing line, and the same rule to determine the end timestamp. 
  

We use the ViTPose~\cite{vitpose}, Hybrid ANN-SNN~\cite{aydin2024hybrid},  LEIR~\cite{relid11d}, and our proposed ResPose for the PMT task. For these methods, the input for ViTPose is RGB frames, and the input for Hybrid ANN-SNN is events. LEIR processes RGB frames and events. We use these HPE methods to perform HPE and then find the shortest time for a specific joint to pass through the line after the starting moment. We calculate the average millisecond time error between the estimated crossing time and the crossing time obtained from events. The experimental results are depicted in Tab.~\ref{tab:timing}. The experimental results verify our expectation that the low-frame-rate methods with RGB input (i.e., ViTPose) fail (error in the order of 50\;$ms$). We would like to point out that a hybrid method (i.e., LEIR) with events and RGB as input, and even a pure event-based method (i.e., Hybrid ANN-SNN) fails on this task too. This suggests that existing HPE methods with low frame-rate input are not suitable for the PMT task. 

In contrast, our ResPose baseline achieves state-of-the-art performance with single-digit millisecond errors (e.g., 4.8\,ms for Punching). This demonstrates that by effectively leveraging high-frequency residual refinement, ResPose can successfully tackle the challenges posed by FlashMotion.

The FlashMotion dataset suggests an abundant opportunity for the HPE community. In addition, we quantitatively evaluate high-speed RGB-based PMT Solutions and reveal that even widely used high-frequency industrial cameras (such as 200Hz) still experience 2-6 ms timing errors during manual labeling of PMT tasks. This error indicates that relying on traditional high-speed photography is not sufficient to achieve true millisecond-level accuracy. Please refer to the Supplementary for details.





\begin{table}[!t]
    \centering
    \caption{High Temporal Resolution HPE Task in FlashMotion.}
    \footnotesize{
    \centering
    \begin{tabular}{lccc}
    \toprule
    Method & MPJPE$\downarrow$ & PCK0.3$\uparrow$ & PCK0.5$\uparrow$ \\
    \midrule
    ViTPose (linear) & 10.06 & 0.96 & 0.98 \\
    ViTPose (spline) & 10.13 & 0.96 & 0.98 \\
    Hybrid ANN-SNN~\cite{aydin2024hybrid} & 22.48 & 0.82 & 0.91 \\
    EventPointPose~\cite{EventPointPose3DV22} & 51.61 & 0.48 & 0.74 \\
    LEIR~\cite{relid11d} & 59.02 & 0.35 & 0.65 \\
    GraphEnet~\cite{GraphEnet} & 43.01 & 0.65 & 0.83 \\
    EvSharp2Blur~\cite{EvSharp} & 8.78 & 0.95 & 0.96 \\
    \midrule
    ResPose (ANN Variant) & 8.12 & 0.95 & 0.96 \\
    \textbf{ResPose (Ours)} & \textbf{5.66} & \textbf{0.97} & \textbf{0.99} \\
    \bottomrule
    \end{tabular}
    }
    \label{tab:hpe}
\end{table}

\subsection{High Temporal Resolution HPE}

\begin{figure}[!tb]
    \centering
     \includegraphics[width=1\linewidth]{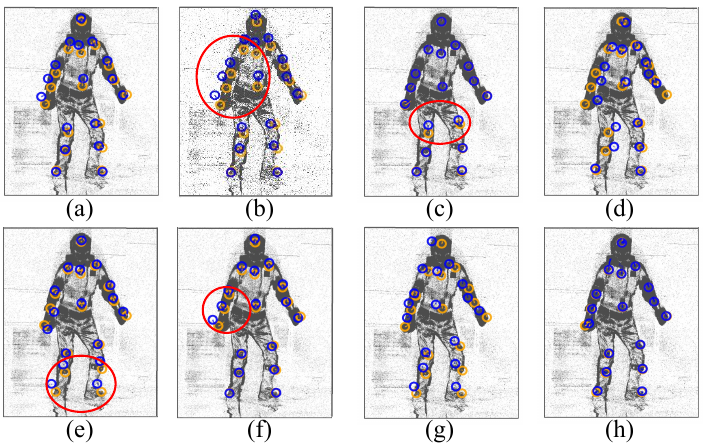}
     \vspace{-4mm}
     \caption{Qualitative comparison of trajectories for the High Temporal Resolution HPE task over a 50\,ms interval. (a) ViTPose. (b) ViT (spline). (c) Hybrid ANN-SNN. (d) EventPointPose. (e) GraphEnet. (f) EvSharp2Blur. (g) LEIR. (h) ResPose (Ours).}
     
     \label{fig:compare}
\end{figure}

We use ViTPose~\cite{vitpose}, Hybrid ANN-SNN~\cite{aydin2024hybrid}, EventPointPose~\cite{EventPointPose3DV22}, GraphEnet~\cite{GraphEnet}, EvSharp2Blur~\cite{EvSharp}, LEIR~\cite{relid11d}, and ResPose for the high temporal resolution HPE task. Due to the insufficient pose output frame rate (below 1000 Hz) of these baselines, we linearly upsample their resulting poses to 1000 Hz. We report the 2D Mean Per Joint Position Error (MPJPE) and the Percentage of Correct Keypoints (PCK0.3, PCK0.5) in Tab.~\ref{tab:hpe}.

ResPose significantly outperforms these baselines, achieving the lowest MPJPE (5.66). This indicates that our method effectively captures micro-movements between frames, establishing a strong benchmark for this task.



Fig.~\ref{fig:compare} visualizes the HPE results over a 50,ms interval. RGB-based ViTPose (a) and its interpolated variant (b) yield large errors for fast blurry motion. Pure event-based methods (c-f) struggle with consistency. Existing fusion baselines like LEIR (g) fail to align with fine-grained motion. In contrast, ResPose (h) generates smooth, high-fidelity trajectories that closely match ground truth dynamics.






\section{Conclusion}

In this work, we present FlashCap, a novel system enabling millisecond-accurate motion capture. Through FlashCap, we collect FlashMotion, the first millisecond-accuracy human motion dataset. 
 We demonstrate its quality through qualitative and quantitative evaluations 
 and show that it exposes the fundamental limitations of existing low-frequency HPE methods. To address these challenges, we proposed ResPose, a simple yet effective Hybrid Spiking-Transformer baseline. The experimental results show that ResPose is promising for high-temporal resolution HPE.
 




\newpage
\section*{Acknowledgments}
This work was partially supported by the Fundamental Research Funds for the Central Universities (No.~20720230033), by Xiaomi Young Talents Program. We would like to thank the anonymous reviewers for their valuable suggestions.

{
    \small
    \bibliographystyle{ieeenat_fullname}
    \bibliography{main} 
}

\end{document}